%% file: main.tex
\PassOptionsToPackage{table,dvipsnames}{xcolor}
\documentclass[10pt,twocolumn,letterpaper]{article}

\usepackage{iccv}

\usepackage{times}
\usepackage{epsfig}
\usepackage{graphicx}
\usepackage{amsmath}
\usepackage{amssymb}
\usepackage{datetime2}
\usepackage{bm}
\usepackage{color,soul}          
\usepackage{microtype}      
\usepackage{dcolumn}        
\usepackage{url}            
\usepackage{booktabs}       
\usepackage{amsfonts}       
\usepackage{nicefrac}       
\usepackage{enumitem}
\usepackage{multirow}
\usepackage{float}
\usepackage{algorithm}
\usepackage[noend]{algpseudocode}
\usepackage[numbers,sort,compress]{natbib}
\usepackage{amsthm}
\usepackage{subfiles}
\usepackage{adjustbox}
\usepackage{subcaption}
\usepackage{makecell}
\usepackage[normalem]{ulem}
\usepackage{caption}
\usepackage{nopageno}
\usepackage[flushmargin,para]{footmisc}
\usepackage[pagebackref,breaklinks,colorlinks]{hyperref} 
\usepackage[table]{xcolor}
\usepackage{pifont}
\newcommand{\cmark}{\ding{51}}%
\newcommand{\xmark}{\ding{55}}%

\usepackage[capitalize]{cleveref}
\crefname{section}{Sec.}{Secs.}
\Crefname{section}{Section}{Sections}
\Crefname{table}{Table}{Tables}
\crefname{table}{Tab.}{Tabs.}
\newcommand{\fig}[1]{Fig.~\ref{#1}}

\definecolor{lightgray}{gray}{0.97}
\definecolor{lightblue}{rgb}{0.93,0.95,1.0}

\iccvfinalcopy 

\def\iccvPaperID{3277} 

\ificcvfinal\fi

\begin{document}

\title{EgoHumans: An Egocentric 3D Multi-Human Benchmark}

\author{
    Rawal Khirodkar, Aayush Bansal, Lingni Ma, Richard Newcombe,  Minh Vo, Kris Kitani \\
    {\urlstyle{sf} \href{https://rawalkhirodkar.github.io/egohumans}{https://rawalkhirodkar.github.io/egohumans}}
}


\twocolumn[{
\maketitle
\ificcvfinal\thispagestyle{empty}\fi
\begin{figure}[H]
\hsize=\textwidth
\centering
\vspace{-8mm}
\includegraphics[width=1.0\textwidth,height=0.27\textwidth]{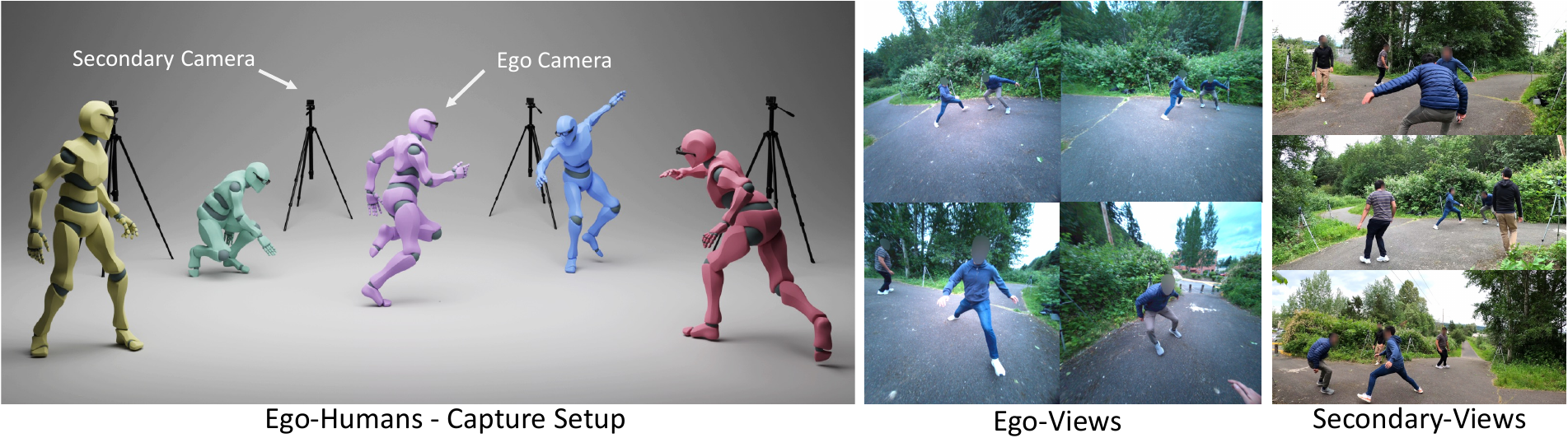}
\vspace*{-0.25in}
\caption{\textcolor{gray}{\textit{Left}}: Our proposed capture setup consists of multiple egocentric cameras from wearable glasses and stationary secondary cameras. This flexible and mobile setup allows us to generate high-quality multi-human 3D annotations for diverse in-the-wild settings.  \textcolor{gray}{\textit{Center}}: Multiple synchronized egocentric views while playing tag. \textcolor{gray}{\textit{Right}}: Synchronized secondary views (cropped) from the stationary cameras. All cameras are spatiotemporally localized in the world coordinate.
}
\label{figure:introduction}
\end{figure}
}]

\begin{abstract}
\vspace{-0.5cm}
We present \textbf{EgoHumans}, a new multi-view multi-human video benchmark to advance the state-of-the-art of egocentric human 3D pose estimation and tracking. Existing egocentric benchmarks either capture single subject or indoor-only scenarios, which limit the generalization of computer vision algorithms for real-world applications. We propose a novel 3D capture setup to construct a comprehensive egocentric multi-human benchmark in the wild with annotations to support diverse tasks such as human detection, tracking, 2D/3D pose estimation, and mesh recovery. We leverage consumer-grade wearable camera-equipped glasses for the egocentric view, which enables us to capture dynamic activities like playing tennis, fencing, volleyball, etc. Furthermore, our multi-view setup generates accurate 3D ground truth even under severe or complete occlusion. The dataset consists of more than 125k egocentric images, spanning diverse scenes with a particular focus on challenging and unchoreographed multi-human activities and fast-moving egocentric views. We rigorously evaluate existing state-of-the-art methods and highlight their limitations in the egocentric scenario, specifically on multi-human tracking. 
To address such limitations, we propose \textbf{EgoFormer}, a novel approach with a multi-stream transformer architecture and explicit 3D spatial reasoning to estimate and track the human pose. EgoFormer significantly outperforms prior art by $13.6$\% IDF1 on the EgoHumans dataset. 
\end{abstract}

\vspace{-0.6cm}
\section{Introduction}
\label{sec:introduction}
\subfile{sections/0_introduction}

\section{Related Works}
\label{sec:related_work}
\subfile{sections/1_related_work}

\section{EgoHumans Dataset}
\label{sec:dataset}
\subfile{sections/2_dataset}

\section{EgoFormer Tracking}
\label{sec:method}
\subfile{sections/3_method}

\section{Experiments}
\label{sec:experiments}

\subfile{sections/4_experiments}

\vspace{-0.2cm}
\section{Discussion}
\label{sec:conclusion}
\subfile{sections/5_conclusion}

\newpage
{\small
\bibliographystyle{ieee_fullname}
\bibliography{references}
}

\end{document}


\title{EgoHumans: An Egocentric 3D Multi-Human Benchmark}

\author{
    Rawal Khirodkar, Aayush Bansal, Lingni Ma, Richard Newcombe, Minh Vo, Kris Kitani
}
\maketitle

\section{Implementation Details}
\label{supp:sec:implementation}
\input{supplementary/sections/0_implementation}

\section{LiDAR Visualization}
\label{supp:sec:lidar}
\input{supplementary/sections/1_lidar}

\section{Other Benchmarks}
\label{supp:sec:benchmark}
\input{supplementary/sections/2_benchmarks}

\section{Qualitative Results}
Please refer to the attached video for qualitative results. We have trimmed videos and compressed the resolution to meet the supplementary submission guidelines - actual resolution and sequence duration follow the main paper.

\newpage
{\small
\bibliographystyle{ieee_fullname}
\bibliography{references}
}

%% file: sections/0_introduction.tex

Understanding humans in 3D from the egocentric view is key to building immersive social telepresence~\cite{lawrence2021project,bamodu2013virtual,ma2021pixel,lombardi2021mixture}, assistive humanoid robots~\cite{goodrich2013teleoperation,fridin2014acceptance,piezzo2017feasibility}, and augmented reality systems~\cite{azuma1997survey,billinghurst2015survey,carmigniani2011augmented}.
A crucial step in this direction is to obtain 3D supervision at scale for deep learning models to generalize to the real world. However, unlike the large-scale 2D benchmarks~\cite{deng2009imagenet,lin2014microsoft,li2019crowdpose,johnson2010clustered,cordts2016cityscapes}, the diversity of the 3D benchmarks~\cite{joo2015panoptic} is severely limited - primarily because manual annotation in the 3D space is impractical. As a result, existing popular 3D benchmarks~\cite{ionescu2013human3,joo2015panoptic,von2018recovering,mehta2017monocular,li2021ai,hassan2019resolving} are constrained to indoor environments or, at most, two human subjects if outdoors, stationary/slow camera motion, with limited occlusion. Furthermore, the majority of these benchmarks only portray the third-person view. Recent progress has been made in constructing egocentric benchmarks~\cite{xu2019mo,ng2020you2me,guzov2021human,zhang2022egobody}. However, they suffer from the same diversity pitfalls, making it difficult to evaluate how close the field is to fully robust and general solutions. To drive advances in the field, we propose a benchmark, \textit{EgoHumans}, that includes challenging scenarios ignored in previous studies and a novel method, \textit{EgoFormer}, that outperforms prior art as a starting point for the evaluations.

\input{tables/0_introduction}

EgoHumans is a new egocentric benchmark consisting of high-resolution videos and comprehensive ground truth annotations such as camera parameters, 2D bounding boxes, human tracking ids~\cite{dendorfer2020mot20}, 2D/3D human poses, and 3D human meshes~\cite{loper2015smpl}. EgoHumans goes beyond previous benchmarks in important ways. First, it captures outdoor videos of unconstrained environments and dynamic human activities, including challenging sporting events such as fencing, badminton, volleyball, etc. Second, the activities are unchoreographed to truly capture the \textit{in-the-wild} philosophy of our work. Our video sequences include fast ego-camera motion, human-human occlusion, truncation, and humans appearing at a wide range of spatial scales. We leverage a flexible multi-camera setup consisting of Meta's Aria glasses~\cite{aria_pilot_dataset}, with an RGB and two greyscale cameras, for the egocentric view and stationary secondary RGB cameras for the auxiliary views (see \fig{figure:sequences}). Such camera combination allows us to accurately track and triangulate human poses in 3D for a long duration without using visual markers~\cite{ionescu2013human3} or additional sensors~\cite{von2018recovering}. The natural form factor of glasses~\cite{maimone2013computational} coupled with the RGB and stereo cameras closely resembles the human vision~\cite{matthies1989dynamic}. Last, as a by-product of our capture setup, we provide 3D annotations for the multi-view secondary cameras. We hope these annotations allow the ability to move fluidly between the egocentric and secondary perspectives~\cite{li2021ego} and inspire new research for holistic human understanding. To our knowledge, EgoHumans is the only multi-human 3D egocentric benchmark with these attributes. 

We generate high-quality 3D ground truth by leveraging state-of-the-art visual-inertial odometry algorithm (VIO)~\cite{aria_pilot_dataset}, which is robust to fast head motion and sudden changes in the eye gaze - frequently observed in natural human behavior~\cite{zhang2020wandering}. All the cameras in our multi-view capture are aligned to a single world coordinate system using Procrustes alignment~\cite{luo2002iterative} of the camera poses. EgoHumans consists of 125k egocentric RGB images and 410k human instance annotations (Tab.~\ref{table:introduction}) capturing high-energy activities in various locations, clothing, and lighting conditions with severe occlusion. We annotate the tracking ids, bounding boxes, and 2D/3D human poses for all views using off-shelf estimators~\cite{jin2020whole,wang2020deep} and manual supervision. With carefully calibrated camera parameters and the multi-view 2D poses for a video, we optimize for 3D skeletons using triangulation~\cite{iskakov2019learnable} and refinement constraints like constant limb length, joint symmetry, and temporal consistency~\cite{vo2020self}. Finally, we build an efficient multi-stage motion capture pipeline to fit the SMPL~\cite{loper2015smpl} body model to the 3D human skeletons.

The scale and diversity of the EgoHumans dataset allow unprecedented opportunities to evaluate and improve egocentric methods. Specifically, we evaluate existing methods for multi-human tracking. Our results show that prior art is susceptible to common failures like person-id switching due to rapid camera motion, occlusion, and unconstrained human activities. Inspired by this, we present \textit{EgoFormer}, a novel 3D human tracking approach with multi-stream transformer architecture that effectively performs human depth reasoning in a camera-agnostic frame of reference. Our proposed method uses self-attention to aggregate multi-view spatial information from the RGB, left, and right stereo cameras simultaneously. EgoFormer significantly outperforms existing state-of-the-art tracking methods~\cite{zhang2022bytetrack,rajasegaran2022tracking,cao2022observation} by $13.6$\% IDF1 score on EgoHumans.

Our contributions are summarized as follows.
\begin{itemize}
\itemsep0em 
  \item \textit{EgoHumans} is the first multi-human 3D egocentric dataset capturing unconstrained human activities in the wild. We provide high-quality 3D ground truth from egocentric and secondary views for all humans. 
  \item We benchmark existing state-of-the-art methods for multi-human tracking and highlight their fundamental limitations on egocentric views. 
  \item We propose \textit{EgoFormer}, a 3D tracking method that uses a multi-stream spatial transformer encoder for depth reasoning from the ego view. Our method consistently outperforms the prior art on the EgoHumans \texttt{test} set.
\end{itemize}

%% file: tables/0_introduction.tex

    
    


\begin{table}[b]
\captionsetup{font=small}
\small

\begin{center}
\vspace*{-0.2in}
\resizebox{3.3in}{!}{
    \setlength{\tabcolsep}{5pt}
    \renewcommand{\arraystretch}{1.2}
    \rowcolors{3}{}{lightgray}
    \begin{tabular}{@{}l|c|c|c|c|c|c|c@{}}
    \Xhline{3\arrayrulewidth}
    
    \textbf{Ego-Datasets} & \textbf{Location} & \textbf{Ego-Views} & \textbf{Sec-Views} & \textbf{Images} & \textbf{Instances}  & \textbf{Mesh} & \textbf{World Co.} \\
    \hline
    Mo2Cap2~\cite{xu2019mo} & indoor & 1 & 0 & 15k & 15k & \xmark & \xmark\\
    You2Me~\cite{ng2020you2me} & indoor & 1 & 0 & 150k & 150k & \xmark & \xmark\\
    HPS~\cite{guzov2021human} & indoor & 1 & 0 & 300k & 320k & \cmark & \cmark\\
    EgoBody~\cite{zhang2022egobody} & indoor & 1 & 5 & 199k & 374k  & \cmark & \cmark\\
    
    \hline
    EgoHumans &  in/outdoor & 4 & 15 & 125k & 410k & \cmark & \cmark\\

    \Xhline{1\arrayrulewidth}
    \end{tabular}
}
\caption{Comparison with 3D ego datasets. \textit{Ego-Views} and \textit{Sec-Views} are number of ego-views and secondary views. \textit{Images} and \textit{Instances} are number of ego-images and self + other visible human instances. \textit{World Co.} refers to world translation and rotation.}
\label{table:introduction}
\vspace*{-0.4in}
\end{center}
\end{table}

%% file: sections/1_related_work.tex
\paragraph{Limited 3D Human Benchmarks.} Throughout the history of computer vision research, benchmarks~\cite{deng2009imagenet,xiao2017fashion,lin2014microsoft,ionescu2013human3,cordts2016cityscapes,geiger2013vision,hoiem2009pascal,li2019crowdpose,zhang2019pose2seg} have played a critical role. However, unlike 2D benchmarks, 3D benchmarks~\cite{ionescu2013human3, von2018recovering, mehta2017monocular,joo2015panoptic,joo2018total,sigal2010humaneva, yi2022mover} are limited in diversity which significantly hampers the ability of deep models to generalize to the real world~\cite{zeng2020srnet}. In addition, 3D human poses are challenging for humans to annotate since the task requires metric precision. As a result, existing datasets rely on wearable sensors~\cite{von2018recovering} or calibrated camera setups~\cite{zhang2022egobody,ionescu2013human3,joo2015panoptic,li2021ai} limited to indoor settings or are entirely synthetic~\cite{patel2021agora,varol2017learning,ros2016synthia,baltieri20113dpes}. Popular datasets like Human3.6M~\cite{ionescu2013human3}, AMASS~\cite{mahmood2019amass}, HumanEva~\cite{sigal2010humaneva}, AIST++~\cite{li2021ai}, HUMBI~\cite{yu2020humbi}, PROX~\cite{hassan2019resolving}, and TotalCapture~\cite{joo2018total} only contain single human sequences. Multi-human datasets like PanopticStudio~\cite{joo2015panoptic}, MuCo-3DHP~\cite{mehta2018single}, TUM Shelf~\cite{chen2020cross} are limited to indoor lab conditions. Outdoor multi-human datasets like 3DPW~\cite{von2018recovering} and MuPoTS~\cite{mehta2018single} have constrained human activities and lack egocentric annotations~\cite{vo2020spatiotemporal,bansal20204d}, or are limited in diversity~\cite{vo2020self}.
Existing egocentric datasets primarily focus on hand-object interactions and action recognition~\cite{bambach2015lending,damen2018scaling,damen2022rescaling,fathi2011understanding,kay2017kinetics,kazakos2019epic,kitani2011fast,kwon2021h2o,li2018eye,narayan2014action,ogaki2012coupling,pirsiavash2012detecting,ryoo2013first,sigurdsson2018actor,yonetani2016recognizing,zhang2022can}. Recent datasets like Mo2Cap2~\cite{xu2019mo}, You2Me~\cite{ng2020you2me}, HPS~\cite{guzov2021human} and EgoBody~\cite{zhang2022egobody} focus on 3D human pose annotations - but are limited to one or two human subjects and indoor settings. We showcase various statistics of EgoHumans against existing ego benchmarks in Tab.\ref{table:introduction} and highlight the key differences.

\vspace{-3mm}
\paragraph{Monocular 3D Human Reconstruction.} Among the recent approaches~\cite{lin2021-mesh-graphormer,kocabas2021pare, lin2021end,li2021hybrik,guan2021bilevel,dwivedi2021learning,zhang2021pymaf,kanazawa2017end, choi2020pose2mesh, khirodkar2022occluded, sun2021monocular}, many rely on the SMPL model~\cite{loper2015smpl}, which offers a low dimensional parametrization of the human body. HMR~\cite{kanazawa2018end} uses a neural network to regress the parameters of an SMPL body from a single image. Follow-up works like SPIN~\cite{kolotouros2019learning}, ROMP~\cite{sun2021monocular}, METRO~\cite{lin2021end}, PARE~\cite{kocabas2021pare}, OCHMR~\cite{khirodkar2022occluded}, SPEC~\cite{kocabas2021spec}, and CLIFF~\cite{li2022cliff} have improved the robustness of the original method in various ways by using additional information like body centers, camera parameters, segmentation mask, 2D pose, etc. Further, methods like VIBE~\cite{kocabas2020vibe}, HMMR~\cite{kanazawa2019learning}, MAED~\cite{wan2021encoder}, and DynaBOA~\cite{guan2022out} predict 3D body parameters from videos. However, most methods require “full-body” images~\cite{pavlakos2022human} and therefore lack robustness when body parts are occluded or truncated, as is often the case in egocentric videos. We also show that existing methods do not exhibit temporal consistency under fast ego-camera motion present in our EgoHumans benchmark.

\vspace{-3mm}
\paragraph{Multi-Object Tracking.} Multi-object tracking is a well-studied area, and we refer the readers to ~\cite{ciaparrone2020deep,dendorfer2021motchallenge,yilmaz2006object,fan2016survey} for a comprehensive summary. In this work, we focus on human tracking methods. Modern tracking methods~\cite{zhang2022bytetrack,bergmann2019tracking,zhang2021fairmot} are primarily driven by bounding-box detections~\cite{ren2015faster,he2017mask,ge2021yolox}, motion models~\cite{bewley2016simple,choi2015near,wojke2017simple,lehmann2006theory}, association algorithms~\cite{mills2007dynamic,grotschel1985solving} or clustering~\cite{vo2020self}. Fundamentally, these methods use 2D representations like body centers~\cite{zhou2020tracking}, keypoints~\cite{raaj2019efficient}, and appearance~\cite{sun2022dancetrack} and lack 3D reasoning crucial to resolving ambiguities posed by severe/complete object occlusion. Recently ~\cite{rajasegaran2022tracking,zhang2022voxeltrack,zou2022snipper} incorporate 3D pose information relative to the camera frame into tracking and report better tracking performance. However, these methods assume stationary/slow camera motion and are unsuitable for rapid ego-camera movement. Our proposed EgoFormer addresses this limitation and performs 3D association in a static global reference frame for tracking.

%% file: sections/2_dataset.tex
\begin{figure*}[t]
\centering
\captionsetup{font=small}
\includegraphics[width=1\linewidth, height=0.26\linewidth]{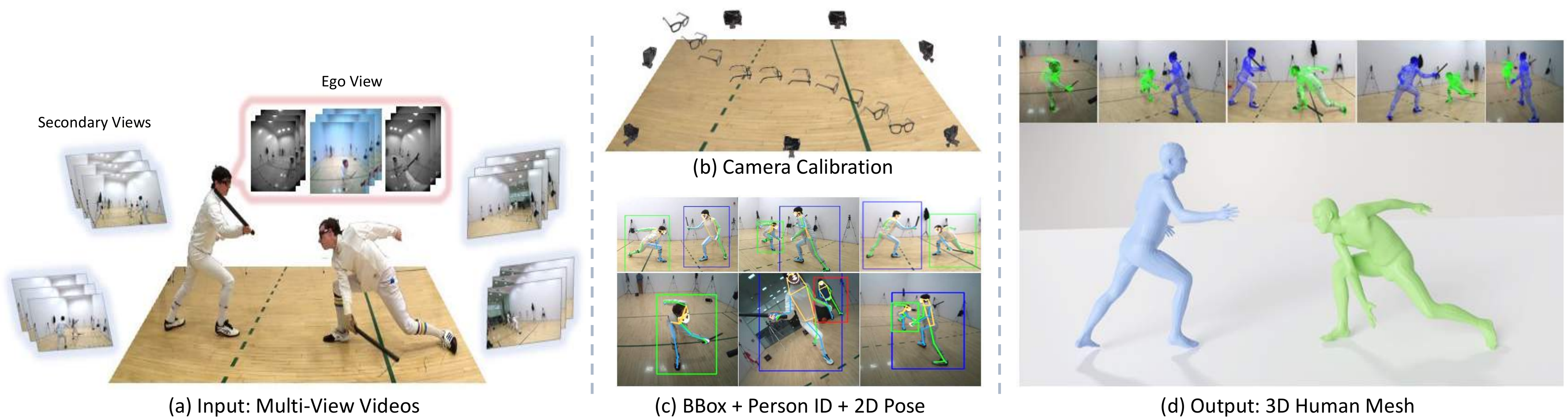}
\vspace*{-0.2in}
\caption{\textbf{Overview of EgoHumans data processing setup.}\textcolor{gray}{(\textit{a})} Multiple synchronized secondary and ego cameras capture the sequence from multiple views.  \textcolor{gray}{(\textit{b})} We align secondary and ego cameras for all time steps into the world coordinate system. \textcolor{gray}{(\textit{c})} For all views, we obtain bboxes, person ids, and 2D poses for all humans. \textcolor{gray}{(\textit{d})} Reconstructed ground-truth meshes overlaid on multiple secondary and ego views.}
\vspace*{-0.2in}
\label{figure:framework}
\end{figure*}

In this section, we describe the data collection setup and annotation algorithms using spatially localized and synchronized multi-view videos. The goal is to design a semi-automatic pipeline to provide ground truth 3D human shapes and poses for egocentric videos. We propose solutions to associate the identities of the subjects consistently across time, compute the body poses, and recover the 3D trajectories of each person in a common frame of coordinates.

\vspace{-3mm}
\paragraph{Data Collection.} For in-the-wild capture, we design a flexible and simple multi-view system with heterogeneous sensors, including multiple Aria glasses and GoPros for the egocentric and secondary views, respectively (c.f., Fig. \ref{figure:framework}a). The glass camera provides a natural human eye's perspective during the capture. The large number of secondary cameras at unique viewpoints ensures robust human pose estimation under occlusions and removes the restriction on the view direction of the subjects. Importantly, the volume created by our cameras is portable and can be moved across locations. Our captures typically consist of $2$ to $6$ egocentric views and $8$ to $15$ secondary views. For Aria glasses, we use $1408 \times 1408$ pixel resolution RGB images and $480 \times 640$ greyscale images. For GoPro cameras, we set the resolution to be $3840 \times 2160$. All cameras are synchronized.

\vspace{-3mm}
\paragraph{Camera Calibration and 3D Localization.} For the egocentric cameras, we obtain the intrinsic parameters of the custom lens from the factory calibration and the per-timestamp extrinsic parameters using state-of-the-art visual-inertial odometry (VIO)~\cite{aria_pilot_dataset}. As the VIO algorithm only provides individual egocentric camera trajectories in an arbitrary coordinate system, we merge multiple egocentric camera trajectories together with the stationary secondary cameras into a single frame of reference by using procrustes-alignment~\cite{luo2002iterative} and structure-from-motion~\cite{schonberger2016structure} (c.f.,Fig. \ref{figure:framework}b).

\vspace{-3mm}
\paragraph{BBox, Identity Association and 2D Human Pose.} In contrast to previous single-human datasets, for our multi-human sequences, solving for consistent person identity throughout the video and across views is a crucial task. We observed that existing state-of-the-art tracking algorithms~\cite{cao2022observation,zhang2022bytetrack,bergmann2019tracking} are prone to failure under occlusion and fast motion. To this end, we obtain an initial 3D region proposal for each subject using the egocentric camera's 3D location and approximating the subject by a 3D cylinder. Further, we obtain per view 2D bounding box (bbox) proposals and ground-truth person ids by reprojecting the 3D proposals (cylinders) to all ego and secondary views. We further refine these 2D bbox proposals using FasterRCNN~\cite{ren2015faster} and manual supervision. Lastly, we annotate the 2D human poses in a top-down fashion for all the views using HRNet-WholeBody~\cite{jin2020whole,wang2020deep} along with manual error fixes (c.f., Fig. \ref{figure:framework}c).

\vspace{-5mm}
\paragraph{3D Human Pose.} Let $C$ be all synchronized video streams from egocentric and secondary cameras with known projection matrices $P_c$. We aim at estimating the global 3D pose $\mathbf{y}_{j,t}\in\mathbb{R}^3$ of a fixed set of human keypoints with indices $j \in (1..J)$ at timestamp $t \in (1..T)$ for all humans in the scene (we omit the human index for simplicity since each subject is processed independently). Let $\mathbf{x}_{c,j,t}\in\mathbb{R}^2$ be the $j$th 2D keypoint at time $t$ from camera $c$. 

To infer the 3D poses from their 2D estimates, we use a linear algebraic multi-view triangulation approach~\cite{hartley2003multiple}. A na\"ive triangulation algorithm assumes that the 2D keypoints $\mathbf{x}_{c,j,t}$ from each view are independent and, thus, make equal contributions to the triangulation. However, in some views, the 2D keypoints cannot be estimated reliably (\eg, due to occlusions or being out of frame), leading to unnecessary degradation of the final triangulation result. This is addressed by applying RANSAC; specifically, for time step $t$, we solve the over-determined system of equations on the homogeneous 3D coordinate vector of the 3D keypoint $\mathbf{\Tilde{y}}_{j,t}$, $A_{j,t}\mathbf{\Tilde{y}}_{j,t} = 0$,
where $A_{j,t} \in \mathbb{R}^{2C' \times 4}$ matrix is composed of the components from the full projection matrices and  $\mathbf{x}_{c,j,t}$ and $C'$ is the cardinality of the camera inlier set after RANSAC. We further refine the per time step 3D pose estimates globally $\mathbf{y}_{\{1..T\}}$ by leveraging human pose priors like constant limb length, joint symmetry, and temporal smoothing~\cite{vo2020self}. Our cost function is given as

\vspace*{-7mm}
\begin{multline}
\mathcal{L}_\text{pose3d}(\mathbf{y}) = w_l\mathcal{L}_\text{limb}(\mathbf{y}) + w_s\mathcal{L}_\text{symm}(\mathbf{y}) \\
+ w_t\mathcal{L}_\text{temporal}(\mathbf{y}) + w_i\mathcal{L}_\text{reg}(\mathbf{y})
\label{eq:fitpose}
\end{multline}
where $\mathbf{y} = \mathbf{y}_{\{1..T\}}$ and $\mathcal{L}_\text{limb}$, $\mathcal{L}_\text{symm}$, $\mathcal{L}_\text{temporal}$, $\mathcal{L}_\text{reg}$ denotes constant limb length, left-right joint symmetry, temporal smoothing, and regularization losses respectively. $w_l, w_s, w_t, w_i$ are scalar weights. Please refer to the supplemental for the loss definitions.

\begin{figure*}[t]
\centering
\captionsetup{font=small}
\includegraphics[width=1\linewidth, height=0.65\linewidth]{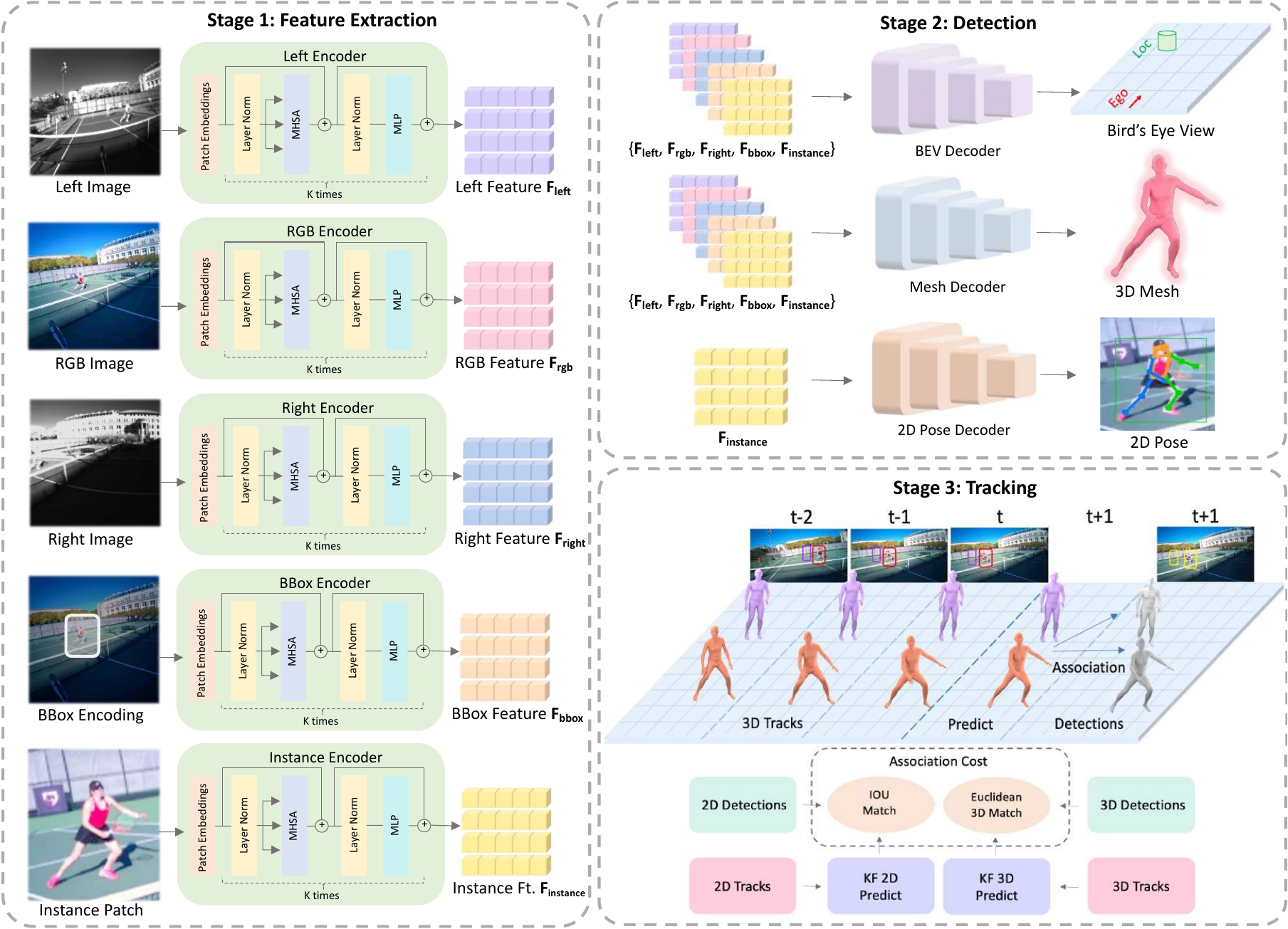}
\vspace*{-0.2in}
\caption{\textbf{Overall architecture of EgoFormer}. Stage 1 extracts multi-view features from the ego images for a human instance. Stage 2 decodes the 2D pose, bird's eye view heatmap of the 3D root location, and the mesh parameters from the features. Stage 3 tracks detections over time steps using two Kalman filters for 2D bounding boxes and 3D root locations.}
\vspace*{-0.2in}
\label{figure:egoformer}
\end{figure*}

\vspace{-3mm}
\paragraph{Mesh recovery.} We represent the human mesh using the body pose and shape, $\boldsymbol{\theta} = [\boldsymbol{\theta_{\text{pose}}}, \boldsymbol{\theta_{\text{shape}}}, \boldsymbol{\theta_{\text{global}}}]$, where $\boldsymbol{\theta_{\text{pose}}} \in \mathbb{R}^{23 \times 6}, \boldsymbol{\theta_{\text{shape}}} \in \mathbb{R}^{10}, \boldsymbol{\theta_{\text{global}}} \in \mathbb{R}^{6}$. The pose parameters $\boldsymbol{\theta_{\text{pose}}}$ are the 6D representation of the joint rotations~\cite{zhou2019continuity} of the $23$ body joints of the SMPL~\cite{loper2015smpl} body. The shape parameters $\boldsymbol{\theta_{\text{shape}}}$ represent the first $10$ coefficients of the PCA shape space learnt from a corpus of registered scans. $\boldsymbol{\theta_{\text{global}}}$ consists of the global root orientation and translation of the body. Similar to ~\cite{huang2017towards,vo2020self}, we fit $\boldsymbol{\theta}$ to the entire 3D pose trajectory in a three-stage optimization scheme. In addition, we use the gender-specific $\boldsymbol{\theta_{\text{shape}}}$ latent space for a better fit to the 3D poses. 

Note, as the mesh fitting procedure is highly under-constrained, obtaining a good initialization for $\boldsymbol{\theta}$ plays an important role in avoiding local minima. To this regard, we run CLIFF~\cite{li2022cliff} on all the camera views and pick the mesh estimate with the lowest joint-reprojection-error (MPJPE)~\cite{von2018recovering} as the initialization. Let $\Phi: \boldsymbol{\theta} \rightarrow \mathbf{y}$ be a differentiable mapping function that projects SMPL parameters $\boldsymbol{\theta}$ to corresponding 3D keypoints $\mathbf{y}$. We define the mesh fitting loss $\mathcal{L}_\text{mesh}(\boldsymbol{\theta})$ as follows,

\vspace*{-0.15in}
\begin{multline}
\mathcal{L}_\text{mesh}(\boldsymbol{\theta}) = w_1|| \mathbf{y} - \Phi(\boldsymbol{\theta})||_2 + w_2|| \boldsymbol{\theta_\text{pose}}||_2 \\
+ w_3\mathcal{L}_\text{limb}\big(\Phi(\boldsymbol{\theta})\big) +  w_4\mathcal{L}_\text{symm}\big(\Phi(\boldsymbol{\theta})\big) \\
+ w_5\mathcal{L}_\text{temporal}\big(\Phi(\boldsymbol{\theta})\big) + w_6\mathcal{L}_{\boldsymbol\beta}(\boldsymbol{\theta}_\text{shape})
\label{eq:mesh}
\end{multline}
where $\mathcal{L}_{\boldsymbol\beta}$ is the Gaussian mixture shape prior loss\cite{bogo2016keep}, the term $|| \boldsymbol{\theta_\text{pose}}||_2$ penalizes hyper-extensions of joints and $w_1.. w_6$ are scalar weights. Other losses are the same as in eq.\ref{eq:fitpose}. We optimize $\mathcal{L}_\text{mesh}$ iteratively in three stages. The first stage consists of optimizing  $\boldsymbol{\theta_{\text{global}}}$, followed by $\boldsymbol{\theta_{\text{shape}}}$ in the second stage and lastly $\boldsymbol{\theta_{\text{pose}}}$ and $\boldsymbol{\theta_{\text{global}}}$ in the third stage. Our recovered meshes are pixel aligned across all egocentric as well as the secondary views (c.f., Fig. \ref{figure:framework}c).

%% file: sections/3_method.tex
In this section, we present EgoFormer -- a simple yet effective multi-stream transformer baseline to track multiple humans from an egocentric camera setup. The goal is to estimate the 3D shape and poses of each observed person over time using the RGB and two greyscale cameras. This task is challenging due to rapid head motion and frequent occlusions. 
The input to EgoFormer is three images, and the output includes the 2D detection, 2D/3D human poses, human shape per person, and identity associations across time. 
We assume the 3D poses of the camera have been reliably estimated from VIO~\cite{zhang2018tutorial}. Fig.~\ref{figure:egoformer} illustrates the three-stage algorithm design, which performs 3D bird's-eye view (BEV) reconstruction in the local camera coordinate and tracks human associations in the global world coordinates. EgoFormer can be trained end-to-end, and we experimentally found that it outperforms modular design variations~\cite{iskakov2019learnable}. We now explain each module in detail.

\vspace{-4mm}
\paragraph{Stage 1 -- Feature Extraction.} As shown in Fig.~\ref{figure:egoformer}, this stage extracts view-dependent features using multiple encoders that share the network architecture. First, the RGB image $I_\text{rgb}$, left/right greyscale images $I_\text{left},I_\text{right} \in \mathbb{R}^{H \times W \times 3}$ are encoded to the corresponding feature maps  $\mathbf{F}_\text{rgb}, \mathbf{F}_\text{left}, \mathbf{F}_\text{right}$, respectively. We resize the images to the same resolution $H \times W$, and the greyscale images are concatenated three times channel-wise to standardize the number of channels. Following ViT~\cite{dosovitskiy2020image}, we first embed the images into tokens via a patch embedding layer. Then the tokens are processed by several transformer layers, where each contains a multi-head self-attention (MHSA) layer and a multi-layer perceptron (MLP) with residual connections. To help the later stage reasoning about the instance, we also perform 2D bbox detection using the off-shelf YOLOX~\cite{ge2021yolox} on the RGB image $I_\text{rgb}$. For each detected bbox, we further obtained two feature encodings. The first $\mathbf{F}_\text{bbox}$ is computed from a boolean representation of the bbox pixels $I_\text{bbox}$. The second $\mathbf{F}_\text{instance}$ is encoded from the cropped patch. Note, all the features $\mathbf{F}$ $\in \mathbb{R}^{\frac{H}{d} \times \frac{W}{d} \times K}$ where $d$ is the downsampling ratio of the patch embeddings (\eg, 16 by default), and $K$ is the channel dimension. 

\begin{figure*}
\centering
\captionsetup{font=small}
\includegraphics[width=1\linewidth]{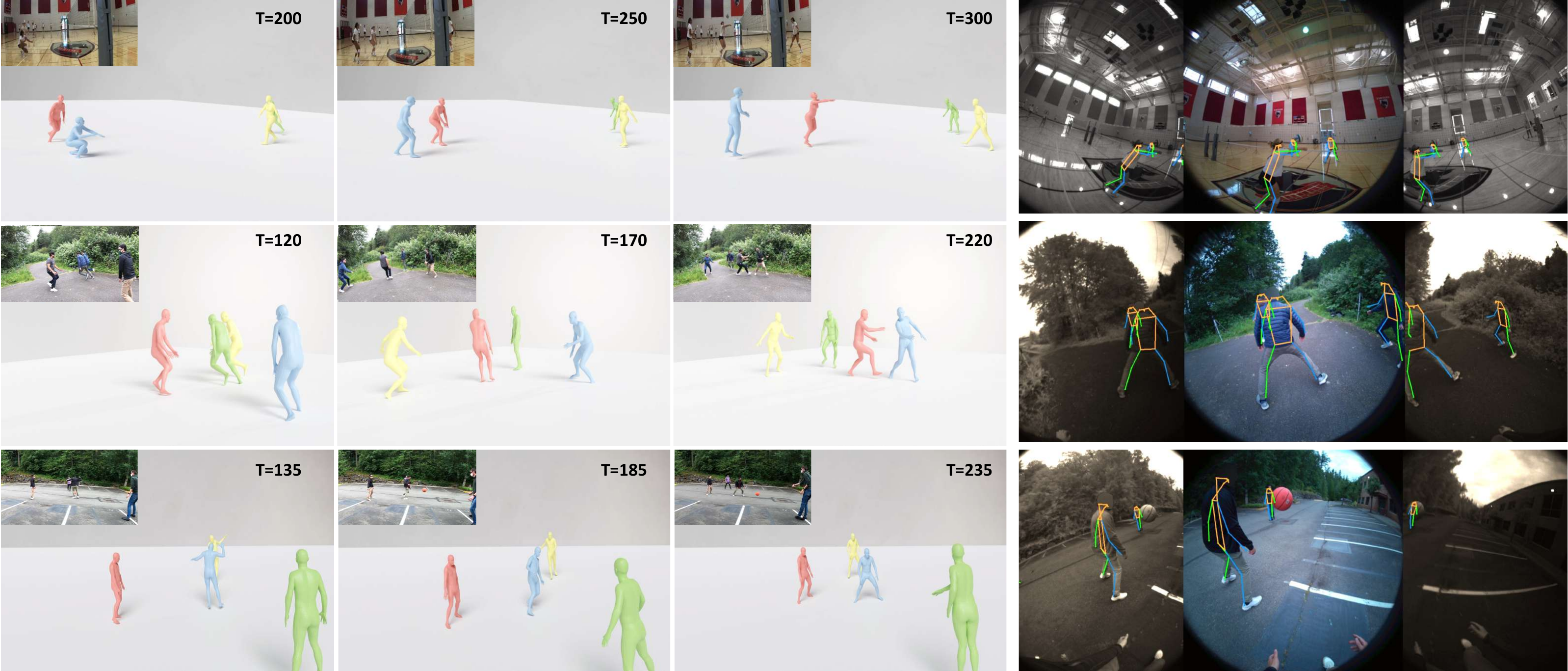}
\vspace*{-0.25in}
\caption{\textbf{Qualitative results for the EgoHumans data processing method.} \textit{Left}: Estimated 3D human meshes with person ids (in color) for playing volleyball and tagging over a long duration of time. \textit{Right:} We visualize the estimated 3D human poses for tennis and badminton by reprojection to the egocentric view. As shown, our multi-view system is robust to occlusion.}
\label{figure:sequences}
\vspace*{-0.2in}
\end{figure*}

\vspace{-5mm}
\paragraph{Stage 2 -- Detection.} We adopt three lightweight decoders to process the extracted features. Every decoder is composed of two deconvolution blocks, where each block contains one deconvolution layer followed by batch normalization~\cite{ioffe2015batch}. The first decoder -- BEV, predicts the target human instance's 3D root location in an \textit{unseen} bird's eye view in the local camera coordinate as a heatmap of $P \times Q$ spatial resolution. We use the log-polar ($\rho, \phi$) to parameterize the root for the BEV heatmap, with a total of $P$ bins for the $\log{\rho}$ and $Q$ bins for the $\phi$. The second decoder is used to regress the 3D SMPL shape and pose $\boldsymbol{\theta}$ using two fully-connected layers. The third pose decoder predicts the 2D keypoints, which regulates the learning by leveraging large-scale 2D annotations from datasets like COCO~\cite{lin2014microsoft}. 

\vspace{-4mm}
\paragraph{Stage 3 -- Tracking.} We design a motion model-based tracker using Kalman Filters (KF)~\cite{Bewley2016_sort}. First, we transform the predicted 3D root locations from the local camera coordinates to the common world reference using the camera poses. Next, we use two filters, KF-2D and KF-3D, to temporally aggregate predictions for the bounding boxes and 3D root locations, respectively. The states $\mathbf{x}_\text{2D} =  [u, v, s, r, \dot{u}, \dot{v}, \dot{s}]$ of KF-2D include the 2D coordinates of the bbox center $(u, v)$, the bbox area $s$ , a constant aspect ratio $r$, and the first derivative $\dot{u}, \dot{v}, \dot{s}$ with respect to time. The states $\mathbf{x}_\text{3D} =  [x, y, z, \dot{x}, \dot{y}, \dot{z}]$ of KF-3D, include the 3D root location $(x, y, z)$ and its velocity $(\dot{x}, \dot{y}, \dot{z})$. The association cost between detection and KF prediction is a weighted sum of $IoU$ match between 2D bboxes and the Euclidean distances between the 3D root locations as illustrated in Fig.~\ref{figure:egoformer}.

\input{tables/1_tracking}

%% file: tables/1_tracking.tex
\begin{table*}[hbt!]
\centering
\small
\captionsetup{font=small}
\resizebox{6.0in}{!}{
\rowcolors{2}{}{lightgray}
\renewcommand{\arraystretch}{1.2} 
\setlength{\tabcolsep}{5pt}
\begin{tabular}{ l | lcc ccccc}
\toprule
Tracker & IDF1$\uparrow$ & MOTA$\uparrow$ & MOTP$\uparrow$ &  FP({\footnotesize $10^4$})$\downarrow$ & FN({\footnotesize $10^4$})$\downarrow$ & IDs$\downarrow$ & Rcll$\uparrow$ & Prcn$\uparrow$  \\
\midrule
SORT~\cite{Bewley2016_sort} & 25.2 & 20.3  & 73.8 & 4.35 & 1.12 & 7,996 & 85.7 & 60.8 \\
DeepSORT~\cite{wojke2017simple} & 38.3 & 22.7  & 73.8 & 4.35 & 1.12 & 6,087 & 85.7 & 60.8 \\
CenterTrack~\cite{zhou2020tracking} & 39.8 & 35.7  & 74.0 & 3.25 & 1.17 & 4,862 & 85.8 & 63.7 \\
FairMOT~\cite{zhang2021fairmot} & 41.2 & 38.0  & 74.1 & 3.73 & 1.21 & 3,915 & 84.5 & 65.1 \\
QDTrack~\cite{pang2021quasi} & 45.5 & 43.8  & 74.3 & 3.11 & 1.21 & 1,074 & 84.6 & 68.2 \\
Tracktor~\cite{bergmann2019tracking} & 43.6 & 55.7  & 71.5 & 2.32 & 1.18 & 1,872 & 83.8 & 66.5\\
PHALP~\cite{rajasegaran2022tracking} & 40.1 & 52.9  & 70.2 & 2.48 & 1.20 & 1,750 & 84.4 & 67.7 \\
OCSORT~\cite{cao2022observation} & 46.4 & 54.6  & 78.9 & 2.44 & 0.82 & 3,486 & 89.2 & 74.4 \\
ByteTrack~\cite{zhang2022bytetrack} & 49.7 & 59.5  & \textbf{78.9} & \textbf{2.10} &  0.82 & 2,696 & 89.5 & \textbf{77.1} \\
\midrule
\textbf{SimpleBaseline (Ours)} & 60.9 (+11.2) & 59.1  & 78.8 & 2.30 & 0.79 & 1,203 & 89.9 & 75.5 \\
\textbf{EgoFormer (Ours)} & \textbf{63.1 (+13.4)} & \textbf{59.8}  & 78.8 & 2.30 & \textbf{0.79} &\textbf{ 741} & \textbf{89.9} & 75.5 \\

\bottomrule
\end{tabular}
}
\caption{Results on the EgoHumans \texttt{test} set. We use the publicly released bounding box detector accompanying the respective methods on MOT17 for evaluation. Our proposed methods SimpleBaseline and EgoFormer, use the same detections as ByteTrack. }
\label{table:tracking}
\vspace*{-0.1in}

\end{table*}

%% file: sections/4_experiments.tex
In this section, we first describe the statistics and quantitative evaluations of the EgoHumans dataset. Then, we perform extensive benchmarking of multi-human tracking algorithms on our dataset.

\subsection{EgoHumans Dataset}

\vspace{-1mm}
\paragraph{Data Statistics.} We collect $7$ sequences from $20$ subjects across $6$ diverse locations (3 indoors and 3 outdoors). The sequences focus on sports activities like basketball, fencing, badminton, tennis, volleyball, tagging, and castle-build (c.f., Fig.~\ref{figure:sequences}). Each sequence has a minimum of 2 and a maximum of 4 subjects wearing Aria glasses and 8 -- 15 secondary cameras, depending on the scenarios. There are $125$k RGB images, $250$k greyscale images from egocentric glasses, and $446$k images from GoPros. We divide each sequence into shorter clips of $30$ seconds on average at $20$ FPS. The annotation per time step includes the calibration and poses per camera, the bounding boxes, person ids, 2D/3D human poses, and 3D shapes per subject. Overall, EgoHumans captures $410$k visible 2D instances. We split the dataset into $77260$ egocentric images for \texttt{train} and $47740$ for \texttt{test}. The split ensures non-overlapping locations between sets.

\vspace{-1mm}
\paragraph{Annotation Accuracy.} We evaluate the end-to-end accuracy by comparing the output 3D human meshes against a dynamic \textit{ground-truth point cloud} obtained by a Kaarta stencil LiDAR~\cite{mate2022evaluation}. We register the point cloud to the scene using ICP~\cite{rusinkiewicz2001efficient} and manual correction. Tab.~\ref{table:lidar} reports the bidirectional Chamfer distance between the recovered 3D human meshes and the point cloud. We analyze the impact of the number of secondary cameras and different losses. As expected, increasing the number of cameras improves annotation. We found $\mathcal{L}_\text{temporal}$ is the most impactful loss.


\subsection{EgoFormer Tracking}

\vspace{-1mm}
\paragraph{Implementation Details.} We follow the common practice to detect instances with YOLOX~\cite{ge2021yolox}, and our network predicts the 3D root location, 2D pose and SMPL parameters per instance. The number of keypoints $J$ is set to 17~\cite{lin2014microsoft}. The encoders are initialized with MAE-Base\cite{MaskedAutoencoders2021} pretrained weights. For the tracking stage, we adopt ByteTrack's association settings. For the tracking baselines, we use their MOT17~\cite{dendorfer2021motchallenge} configuration as default. The input resolution is set to $256 \times 192$. The model is trained for $210$ epochs with AdamW~\cite{reddi2019convergence} with $5e-4$ learning rate,  decayed by $10$ at the $170$th and $200$th epoch on $8$ A6000 GPUs on a combination of EgoHumans and COCO dataset. Note only \textit{feature-extraction} and \textit{detection} stages of EgoFormer have learnable parameters.

\input{tables/2_lidar}

\vspace{-3mm}
\paragraph{Metrics.} To evaluate the 3D human tracking performance, we use the CLEAR metrics~\cite{bernardin2008evaluating}, including MOTA, FP, FN, IDs, \etc along with IDF1\cite{ristani2016performance} and HOTA~\cite{jonathon2021hota}. MOTA focuses on bbox detection accuracy. IDF1 evaluates the instance identity preservation and focuses on the association performance. Recently, HOTA has been proposed, which explicitly balances the effect of accurate detection and consistent association. Our experiments predominantly use an off-shelf bbox detector, so IDF1 is our primary metric.

\vspace{-3mm}
\paragraph{Baselines.} We benchmark a wide set of algorithms on EgoHumans to analyze the state-of-the-art performance for egocentric multi-human tracking. These algorithms include SORT~\cite{Bewley2016_sort}, DeepSORT~\cite{wojke2017simple}, CenterTrack~\cite{zhou2020tracking}, FairMOT~\cite{zhang2021fairmot}, QDTrack~\cite{pang2021quasi}, Tracktor~\cite{bergmann2019tracking}, PHALP~\cite{rajasegaran2022tracking}, OCSORT~\cite{cao2022observation} and ByteTrack~\cite{zhang2022bytetrack}. For fair analysis, we used the published models. Since these algorithms only consider single RGB input without training from our data, we design a baseline -- \textit{SimpleBaseline} to be directly comparable.
Specifically, we use the monocular depth estimator MiDaS~\cite{ranftl2020towards} to predict a dense depth map for the RGB image. The 3D root location of each detection is obtained by averaging the depth of pixels inside the bbox. From here, the SimpleBaseline shares the same \textit{tracking} stage as the EgoFormer, where the 3D root locations in the camera coordinates are transformed to the world coordinate via the camera poses to compute the KF-3D's state. Note the SimpleBaseline does not need to be trained on the EgoHumans dataset since the tracking stage contains no network parameters.

\input{tables/3_finetuned_tracking}

\begin{figure*}
\centering
\captionsetup{font=small} 
\includegraphics[width=0.9\linewidth,height=0.38\linewidth]{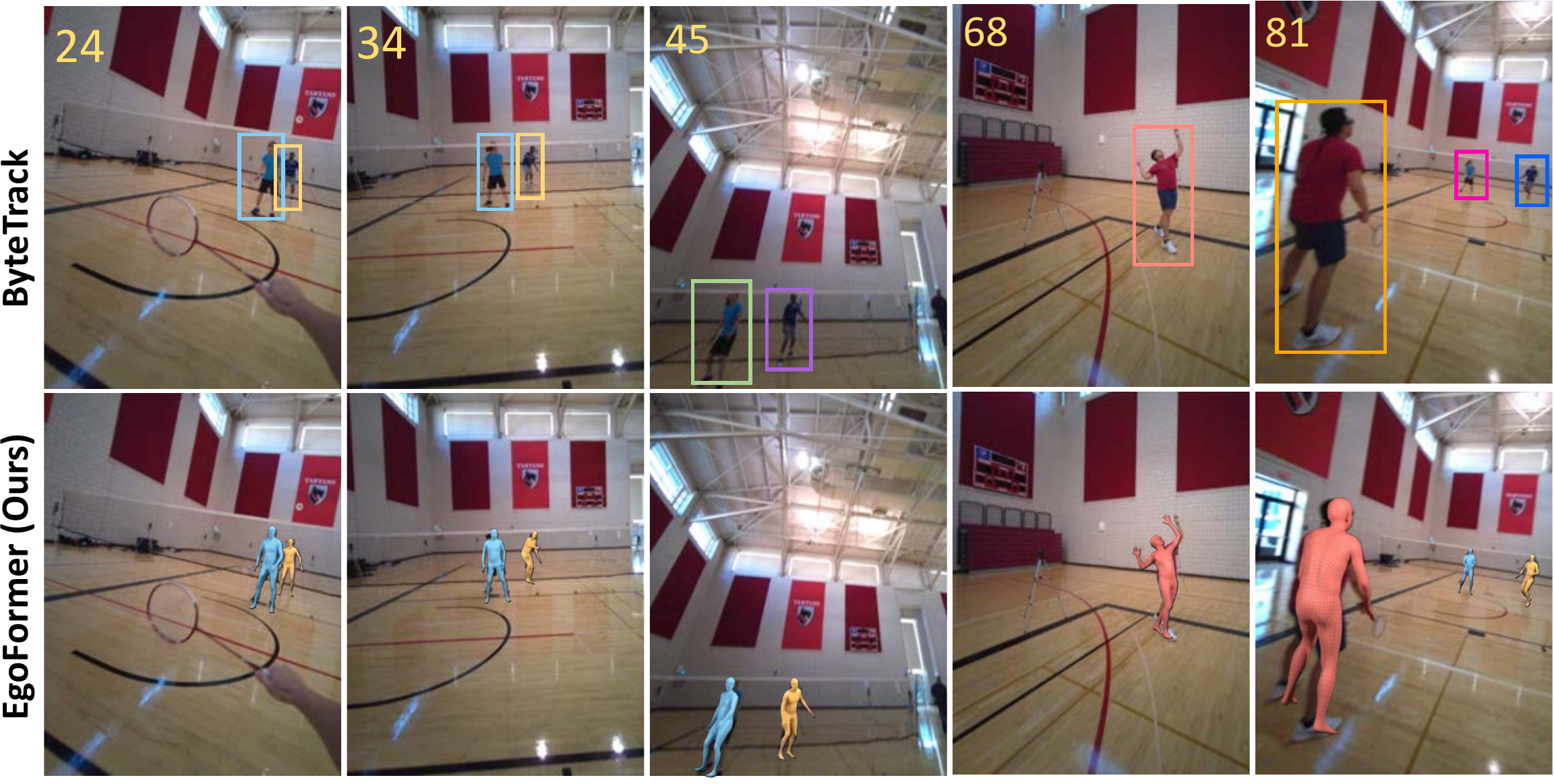}
\includegraphics[width=0.9\linewidth,height=0.38\linewidth]{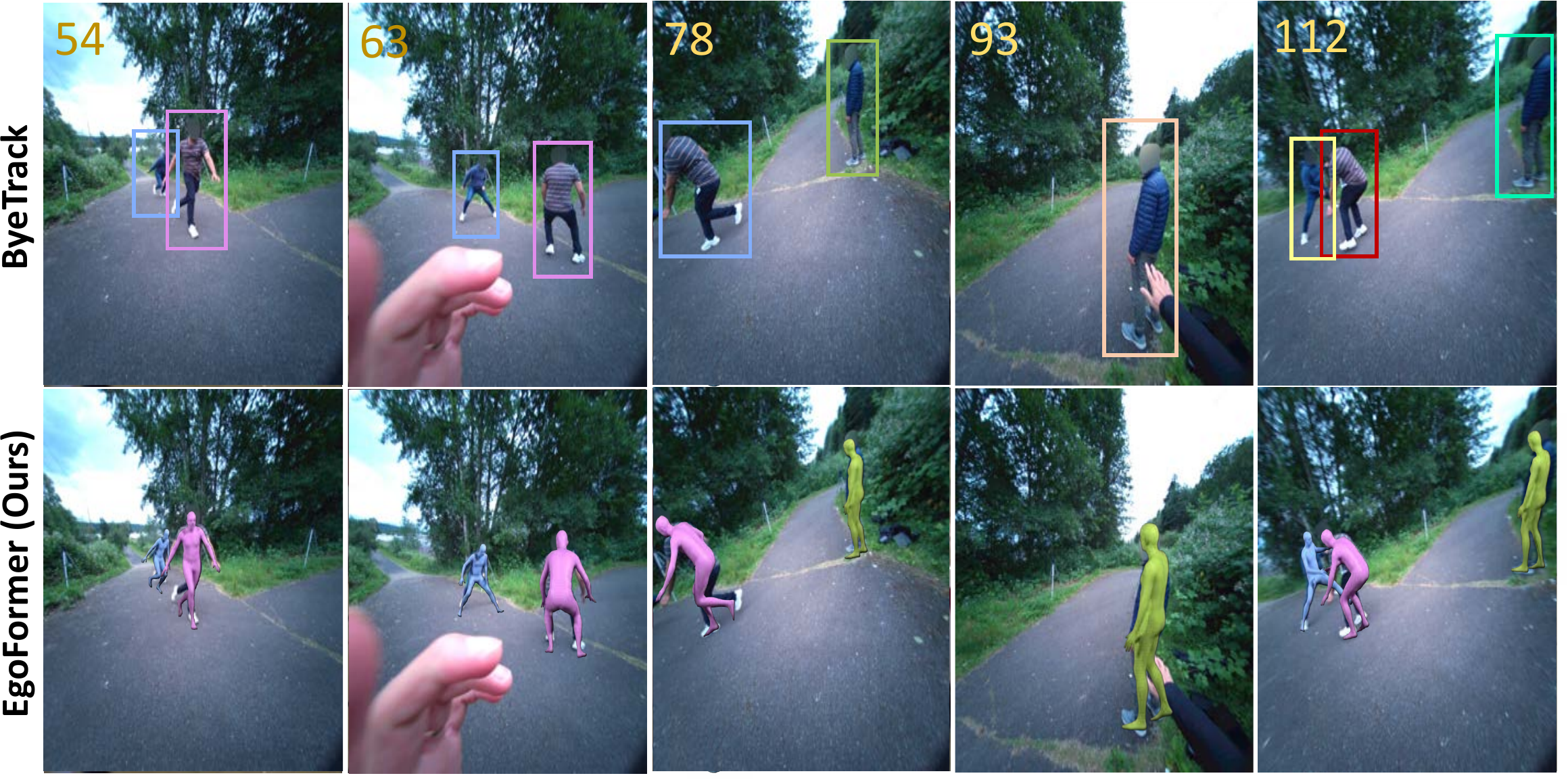}

\vspace*{-0.1in}
\caption{\textbf{Qualitative results of EgoFormer for tracking}. In comparison to a 2D approach like ByteTrack~\cite{zhang2022bytetrack}, our proposed method performs identity association in 3D and suffers from fewer identity switches. Predicted person ids are shown in color. }
\label{figure:qualitative}
\vspace*{-0.1in}
\end{figure*}

\vspace{-3mm}
\paragraph{Results.} Tab.~\ref{table:tracking} compares the performance of EgoFormer, SimpleBaseline, and other methods on the EgoHumans \texttt{test} set. EgoFormer significantly outperforms ByteTrack by $13.4$\% IDF1 highlighting the superior instance association. We observe that MOTA is comparable since the same detections are used. Our method also drastically reduces the identity switches to $741$ compared to the prior art. EgoFormer implicitly learns to solve for person-id across views leveraging a larger field of view. We observe similar performance gains for SimpleBaseline over previous methods showcasing that tracking with the 3D association is crucial for the egocentric view. We show the qualitative tracking results in Fig.~\ref{figure:qualitative}.

\vspace{-3mm}
\paragraph{Finetuned Baselines.} To evaluate the effectiveness of our \texttt{train} set, we finetune tracking baselines on EgoHumans. We use the same hyperparameters provided by the authors for the MOT17 dataset for all the methods and combine MOT17 and EgoHumans \texttt{train} sets for training. As shown in Tab.~\ref{table:finetuned_tracking}, we observe the average gain of $2.1\%$ IDF1 for all methods. However, baselines are still limited by their 2D nature, hence unsuitable for egocentric reasoning.

\vspace{-3mm}
\paragraph{Ablations.} Tab.~\ref{table:ablation_input} compares the performance of EgoFormer with a varying set of input images and decoders. Using all three images gives the best results due to a wider \textit{field-of-view} aiding in better depth-reasoning. In addition, we find the 2D pose decoder with COCO annotations crucial for generalization to unseen scenes.

\input{tables/4_ablation_input}

%% file: tables/2_lidar.tex
\begin{table}[b]
\centering
\small
\resizebox{3.0in}{!}{
    \renewcommand{\arraystretch}{1.2}
    \rowcolors{1}{}{lightgray}
    \setlength{\tabcolsep}{4pt}
  
    \begin{tabular}{@{}c|c|c|c|c|c @{}}
        \Xhline{3\arrayrulewidth}
       Cameras & $\mathcal{L}_\text{limb}$ & $\mathcal{L}_\text{symm}$ & $\mathcal{L}_\text{temporal}$ & $\mathcal{L}_\text{reg}$ & Error (cm) \\
        \hline
        4 & \cmark &\cmark &\cmark &\cmark   & $10.4$   \\ 
        8 & \cmark &\cmark &\cmark &\cmark   & $8.3$  \\ 
        12 &\xmark &\cmark &\cmark &\cmark   & $6.6$  \\
        12 &\cmark &\xmark &\cmark &\cmark   & $7.1$  \\
        12 &\cmark &\cmark &\xmark &\cmark   & $7.9$  \\
        12 &\cmark &\cmark &\cmark &\xmark   & $7.6$  \\
        12 &\cmark &\cmark &\cmark &\cmark   & $\textbf{5.8}$  \\ 
         \Xhline{3\arrayrulewidth}
    \end{tabular}
}
\caption{3D error (in cm) of our localization with a varying number of secondary cameras and 3D pose refinement losses. Increasing the number of cameras reduces the error due to better coverage. The relative importance of $\mathcal{L}_\text{temporal}$ is greater than $\mathcal{L}_\text{limb}$ and $\mathcal{L}_\text{symm}$.}
\label{table:lidar}
\vspace*{-0.3in}
\end{table}

%% file: tables/3_finetuned_tracking.tex
\begin{table}[b]
    \vspace*{-0.2in}
    \centering
    \small
    \resizebox{3.0in}{!}{
        \setlength\tabcolsep{4pt}
        \renewcommand{\arraystretch}{1.2} 
        \rowcolors{1}{}{lightgray}
        \begin{tabular}{@{}l|c c c l}
       \toprule
        Tracker & IDF1$\uparrow$ &  HOTA$\uparrow$ & MOTA$\uparrow$ & IDs$\downarrow$\\
        \midrule
        DeepSORT~\cite{wojke2017simple} & 40.1 & 30.5  & 23.6 & 5,971  \\
        QDTrack~\cite{pang2021quasi} & 46.1 & 34.1  & 44.6 & 1,342  \\
        Tracktor~\cite{bergmann2019tracking} & 42.8 & 33.4  & 53.2 & 3,312  \\
        OCSORT~\cite{cao2022observation} & 47.2 & 37.9  & 56.1 & 2,430  \\
        ByteTrack~\cite{zhang2022bytetrack} & 51.5 & 40.6  & 59.7 & 1,203  \\
        \textbf{EgoFormer (Ours)} & \textbf{63.1} & \textbf{48.1}  & \textbf{59.8} & \textbf{741}  \\
        
            \bottomrule
        \end{tabular}
    }

    \caption{Comparison of EgoFormer with state-of-the-art tracking baselines. All methods are fine-tuned on the EgoHumans \texttt{train} set.}
    \label{table:finetuned_tracking}
\end{table}

%% file: tables/4_ablation_input.tex
\begin{table}
    \centering
    \small
    \resizebox{3.3in}{!}{
        \setlength\tabcolsep{6pt}
        \renewcommand{\arraystretch}{1.2} 
        \rowcolors{1}{}{lightgray}
        \begin{tabular}{@{}l| c c c l}
       \toprule
        Method & IDF1$\uparrow$ &  HOTA$\uparrow$ & MOTA$\uparrow$ & IDs$\downarrow$\\
        \midrule
        RGB, \scriptsize{\textit{FOV =} $110^{\circ}$}  & 58.1 & 40.5  & 59.5 & 2,139  \\
        RGB + Left, \scriptsize{\textit{FOV =} $130^{\circ}$} & 61.6 & 45.8  & 59.5 & 1,102  \\
        RGB + Right, \scriptsize{\textit{FOV =} $130^{\circ}$} & 61.2 & 45.2  & 59.4 & 1,164  \\
        \midrule
        w/o 2D Pose Decoder & 56.8 & 38.2  & 59.4 & 2,877  \\
        w/o 3D Mesh Decoder & 62.5 & 46.8  & 59.6 & 983  \\
        Full system, \scriptsize{\textit{FOV =} $150^{\circ}$} & \textbf{63.1} & \textbf{48.1}  & \textbf{59.8} & \textbf{741}  \\
            \bottomrule
        \end{tabular}
    }
    \caption{\textbf{Ablation of the main components of EgoFormer}. We observe that increasing field-of-view and regularization using the 2D pose decoder reduces identity switches.}
    \label{table:ablation_input}
    \vspace*{-0.3in}
\end{table}

%% file: sections/5_conclusion.tex
We introduced a new in-the-wild 3D benchmark for detection, tracking, pose estimation, and mesh recovery of humans from egocentric captures. Emphasis was placed on capturing unchoreographed, dynamic activities in the real world. Our evaluations show that existing state-of-the-art methods are not suited for rapid camera motion present in wearable ego cameras. We believe that EgoHumans is a significant conceptual change for 3D datasets and will inspire a new research direction for egocentric methods. We also present EgoFormer, a simple 3D human tracker with multi-stream transformer architecture and explicit 3D spatial reasoning which outperforms existing methods by a significant margin.

\vspace{1mm}\noindent 
\textbf{Limitations.} At present, we trade off 3D keypoint localization accuracy in favor of an in-the-wild capture. There is still a particular gap between the performance of static indoor wired 3D capture systems and our capture setup due to errors in camera synchronization and calibration.

%% file: supplementary/sections/0_implementation.tex
In this section, we provide additional implementation details about EgoHumans and EgoFormer.

\subsection{EgoHumans 3D Loss Functions}
We define the 3D pose losses $\mathcal{L}_\text{limb}$, $\mathcal{L}_\text{symm}$ and $\mathcal{L}_\text{temporal}$.
Assume $\mathbf{y} = \mathbf{y}_{\{1..T\}}$ where $\mathbf{y}_t \in \mathbb{R}^{J \times 3}$ is the 3D coordinates of $J$ joints at time $t$. Let $\Lambda \in J \times J$ be the set of start and end joint indexes representing human limbs, \eg left and right-leg, left and right-hand \etc. Thus, the limb lengths at time $t$, $\boldsymbol{\lambda}_t = \{||\mathbf{y}^\text{start}_t-\mathbf{y}^\text{end}_t|| : \text{(start, end)} \in \Lambda\}$. Further, let $\boldsymbol{\lambda}^\text{left}_t$ and $\boldsymbol{\lambda}^\text{right}_t$ be all the left and right limb lengths at time $t$. Then,

\begin{eqnarray}
   \mathcal{L}_\text{limb}(\mathbf{y}) & = &\sum_{t=1}^{T-1} || \boldsymbol{\lambda}_{t+1} - \boldsymbol{\lambda}_t||\\
    \mathcal{L}_\text{symm}(\mathbf{y}) & = &\sum_{t=1}^{T} || \boldsymbol{\lambda}^\text{left}_{t} - \boldsymbol{\lambda}^\text{right}_{t}||\\
    \mathcal{L}_\text{temporal}(\mathbf{y}) & = &\sum_{t=1}^{T-1} || \mathbf{y}_{t+1} - \mathbf{y}_t||
\end{eqnarray}

\subsection{EgoFormer Objective Function}
The objective function for EgoFormer training is a weighted sum of each decoder's individual loss $\mathcal{L}_\text{bev}$, $\mathcal{L}_\text{mesh}$, and $\mathcal{L}_\text{pose2d}$.
Let $I_\text{rgb}$, $I_\text{left}, I_\text{right} \in \mathbb{R}^{H \times W \times 3}$ be the input images and $I_\text{bbox}$, $I_\text{instances}$ be the bounding box encoding and the instance patch respectively. We represent the overall input $\{I_\text{rgb}, I_\text{left}, I_\text{right}, I_\text{bbox}, I_\text{instances}\}$ as $X$. 

\begin{eqnarray}
    \hat{y}_\text{bev} &  = &\mathtt{BEV}(\mathbf{F}_\text{left} + \mathbf{F}_\text{rgb} + \mathbf{F}_\text{right} + \mathbf{F}_\text{bbox} + \mathbf{F}_\text{instance}) \\
     \hat{y}_\text{theta} &  = &\mathtt{Mesh}(\mathbf{F}_\text{left} + \mathbf{F}_\text{rgb} + \mathbf{F}_\text{right} + \mathbf{F}_\text{bbox} + \mathbf{F}_\text{instance}) \\
     \hat{y}_\text{pose2d} &  = &\mathtt{Pose2D}( \mathbf{F}_\text{instance}) \\
\end{eqnarray}

Let $y_\text{bev}$ be the target BEV heatmap, $y_\text{theta}$ be the target mesh parameters $\boldsymbol\theta$ and  $y_\text{pose2d}$ be the multi-channel target 2D pose heatmap, 

\begin{eqnarray}
    \mathcal{L}_\text{bev}(X) &  = &\mathtt{MSE}(\hat{y}_\text{bev}, y_\text{bev}) \\
    \mathcal{L}_\text{mesh}(X) & = &||\hat{y}_\text{theta} - y_\text{theta}|| \\
    \mathcal{L}_\text{pose2d}(X) & = &\mathtt{MSE}(\hat{y}_\text{pose2d}, y_\text{pose2d})
\end{eqnarray}

%% file: supplementary/sections/1_lidar.tex
Our LiDAR evaluation setup consists of $15$ secondary views and $2$ ego views. Two subjects \textit{target} and \textit{observer} participate in the sequence where we report the 3D error for the \textit{target} subject only. Please refer to the main paper for the evaluations.

\begin{figure}[H]
\centering
\includegraphics[width=1\linewidth,height=1.1\linewidth]{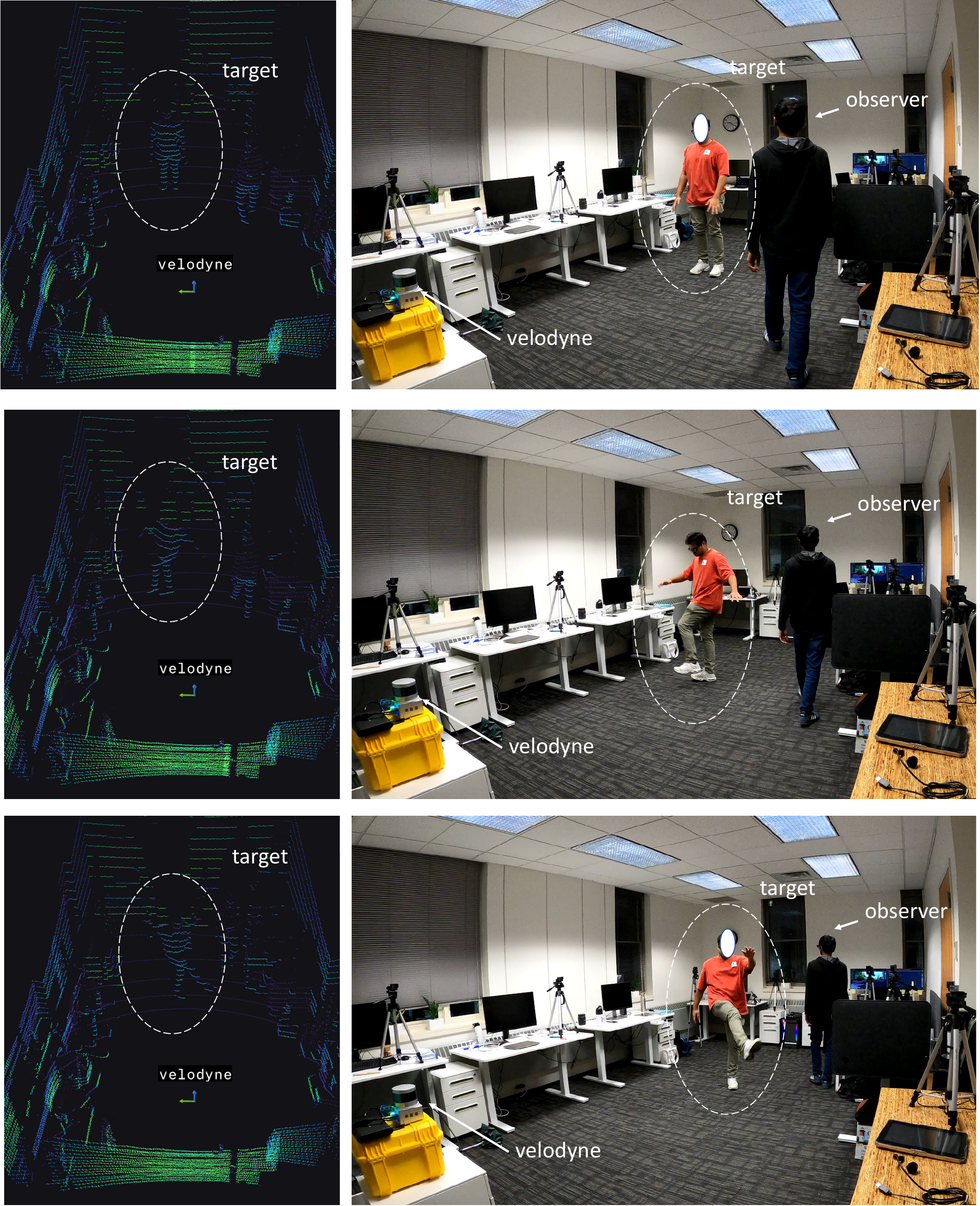}
\caption{\textbf{LiDAR Visualization}. \textit{Left:} Velodyne LiDAR  ground-truth point cloud for the target subject.\textit{Right:} Secondary view of the scene with the target and observer with $15$ secondary views. Both the subjects have wearable Aria glasses.}
\label{supp:figure:lidar}
\end{figure}

%% file: supplementary/sections/2_benchmarks.tex
For completeness, we provide evaluations for 2D bounding box detection, 2D pose estimation, and 3D mesh reconstruction using state-of-the-art methods.

\subsection{Bounding Box Detection}
We compare the state-of-the-art 2D bounding box detectors on the EgoHumans \texttt{test} set in Tab.\ref{supp:table:bbox}. All the detectors are trained on the COCO~\cite{lin2014microsoft} dataset. We use the same evaluation metrics as the COCO dataset. Lastly, we observe the best performance of $41.1$ AP by YOLOX with a large backbone showcasing that our annotations are challenging.

\input{supplementary/tables/0_bbox}

\subsection{2D Pose Estimation}
We compare the state-of-the-art 2D pose estimators on the EgoHumans \texttt{test} set in Tab.\ref{supp:table:pose2d}. All the models are top-down in nature and are trained on the COCO~\cite{lin2014microsoft} dataset. For simplicity, we use ground-truth bounding boxes for evaluations. We observe the best performance of $78.7$ AP by HRNet~\cite{he2016deep} with UDP~\cite{huang2020devil} post-processing and an HRNet-W48 backbone.

\input{supplementary/tables/1_pose2d}

\newpage
\subsection{Human Mesh Reconstruction}
To quantitatively evaluate the performance of the off-shelf mesh reconstruction methods on the EgoHumans \texttt{test} set, we report PVE, MPJPE, PA-PVE, and PA-MPJPE as our primary evaluation metrics in Tab.\ref{supp:table:mesh}. They are all reported in millimeters (mm) by default. Among these metrics, PVE denotes the mean Per-vertex Error, defined as the average point-to-point Euclidean distance between the predicted and ground truth mesh vertices, while MPJPE denotes the Mean Per Joint Position Error. PA-PVE and PAMPJPE denote the PVE and MPJPE after rigid alignment of the prediction with the ground truth using Procrustes Analysis. Note that the metrics PA-PVE and PA-MPJPE are not aware of the global rotation and scale errors since they are calculated after rigid alignment.

\input{supplementary/tables/2_mesh}

%% file: supplementary/tables/0_bbox.tex
\begin{table*}[h]
    \centering
    \renewcommand{\arraystretch}{1.2} 
    \setlength{\tabcolsep}{7pt}
    \begin{tabular}{@{}l|c|c c c c c c c c c@{}}
    \toprule

Method  & Backbone  & $\text{AP}$ & $\text{AP}^{50}$ & $\text{AP}^{75}$ & $\text{AP}^\text{L}$ & $\text{AR}$ & $\text{AR}^\text{M}$ & $\text{AR}^\text{L}$  \\
    \midrule
    
      Swin Transformer~\cite{liu2021swin}  & Tiny & 37.6  & 76.4 & 27.4 & 38.4 & 51.2 & 38.5 & 51.3  \\
      Swin Transformer~\cite{liu2021swin}  & Small & 36.3  & 73.5 & 26.7 & 37.1 & 51.4 & 38.9 & 51.6  \\
      DETR~\cite{carion2020end} & ResNet-50 & 35.6  & 75.8 & 21.9 & 36.4 & 50.0 & 32.4 & 50.2  \\
      Deformable DETR~\cite{zhu2020deformable} & ResNet-50 & 38.8   & 78.1 & 30.3 & 39.6 & 52.5 & \textbf{41.8} & 52.6  \\
      FasterRCNN~\cite{ren2015faster} & ResNet-50 & 39.2 & 78.7 & 30.4 & 39.6 & 51.4 & 37.0 & 51.6  \\
      FasterRCNN~\cite{ren2015faster} & ResNet-101 & 39.4 & 77.6 & 32.5 & 39.8 & 51.6 & 37.4 & 51.8  \\
      YOLOX~\cite{ge2021yolox} & Tiny & 34.8 & 72.9 & 24.2 & 35.4 & 48.0 & 33.3 & 48.2 \\
      YOLOX~\cite{ge2021yolox} & Small & 38.5 & 77.0 & 29.7 & 39.0 & 51.2 & 36.8 & 51.4 \\
      YOLOX~\cite{ge2021yolox} & Large & \textbf{41.1} & \textbf{79.0} & \textbf{37.3} & \textbf{41.6} & \textbf{53.7} & 41.4 & \textbf{53.8} \\
      
\bottomrule
  \end{tabular}

    \caption{Comparison of off-shelf bounding box detectors on the EgoHumans \texttt{test} set trained on the COCO dataset.}
    \label{supp:table:bbox}
\end{table*}

%% file: supplementary/tables/1_pose2d.tex
\begin{table*}[h]
    \centering
    \renewcommand{\arraystretch}{1.2} 
    \setlength{\tabcolsep}{7pt}
    \begin{tabular}{@{}l|c|c c c c c |c c c c c@{}}
    \toprule

Method  & Backbone  & $\text{AP}$ & $\text{AP}^{50}$ & $\text{AP}^{75}$ & $\text{AP}^\text{M}$ & $\text{AP}^\text{L}$ & $\text{AR}$ & $\text{AR}^{50}$ & $\text{AR}^{75}$ & $\text{AR}^\text{M}$ & $\text{AR}^\text{L}$  \\
    \midrule
    
      SimpleBaseline~\cite{xiao2018simple}  & ResNet-50 & 74.6  & 93.9 & 86.3 & 53.9 & 74.9 & 78.0 & 94.7 & 87.8 & 61.7 & 78.1  \\
      SimpleBaseline~\cite{xiao2018simple}  & ResNet-101 & 75.2  & 93.9 & 87.2 & 54.0 & 75.3 & 78.7 & 94.6 & 88.3 & 62.3 & 78.9  \\
      SimpleBaseline~\cite{xiao2018simple}  & ResNet-152 & 75.5  & 93.9 & 87.3 & 56.1 & 75.7 & 78.9 & 94.8 & 88.8 & 63.4 & 79.1  \\
      Swin Transformer~\cite{liu2021swin}  & Tiny & 72.6 & 93.9 & 86.0 & 53.8 & 72.8 & 76.8 & 94.7 & 87.7 & 63.5 & 76.9  \\
      Swin Transformer~\cite{liu2021swin}  & Base & 75.0 & 93.9 & 87.2 & 53.2 & 75.2 & 78.5 & 94.8 & 88.6 & 62.5 & 78.7  \\
      Swin Transformer~\cite{liu2021swin}  & Large & 76.6  & 94.9 & 88.2 & 54.6 & 76.8 & 79.9 & 95.2 & 89.6 & 62.8 & 80.1  \\
      HRFormer~\cite{yuan2021hrformer}  & Small & 76.3  & 94.9 & 88.2 & 54.6 & 76.7 & 79.7 & 95.1 & 89.0 & 65.0 & 79.8  \\
      HRFormer~\cite{yuan2021hrformer}  & Base & 76.2  & 94.9 & 88.3 & 54.9 & 76.5 & 80.1 & 95.3 & 89.9 & 65.4 & 80.2  \\
      HRNet-UDP~\cite{he2016deep,huang2020devil}  & HRNet-W32 & 76.4  & 94.9 & 88.4 & 57.0 & 76.7 & 79.9 & 95.2 & 89.6 & 66.5 & 80.0  \\
      HRNet-UDP~\cite{he2016deep,huang2020devil}  & HRNet-W48 & \textbf{78.7}  &\textbf{ 94.9} & \textbf{89.4} & \textbf{57.0} & \textbf{78.8} & \textbf{81.6} & \textbf{95.3} & \textbf{90.2} & \textbf{66.8} & \textbf{81.8}  \\
\bottomrule
  \end{tabular}

    \caption{Comparison of off-shelf top-down 2D pose estimators trained on the COCO dataset evaluated on the EgoHumans \texttt{test}. We use ground-truth bounding boxes at an input resolution of $256 \times 192$ for evaluation.}
    \label{supp:table:pose2d}
\end{table*}

%% file: supplementary/tables/2_mesh.tex
\begin{table}[h]
    \centering
        \renewcommand{\arraystretch}{1.2} 
        \setlength{\tabcolsep}{7pt}
        \begin{tabular}{@{}l|c c |c c}
        \toprule
        Method  & MPJPE$\downarrow$ & PVE$\downarrow$ & PA-MPJPE$\downarrow$ &  PA-PVE$\downarrow$\\
        \midrule
        HMR~\cite{kanazawa2018end} & 127.4 & 154.9 & 84.5 & 121.0 \\
        SPIN~\cite{kolotouros2019learning} & 120.3 & 147.3 & 82.1 & 118.5 \\
        VIBE~\cite{kocabas2020vibe} & 111.8 & 128.1 & 78.9 & 110.4 \\
        PyMaf~\cite{zhang2021pymaf} & 96.8 & 109.5 & 77.5 & 108.7 \\
        OCHMR~\cite{khirodkar2022occluded} & 94.2 & 108.9 & 75.0 & 102.3 \\
        ROMP~\cite{sun2021monocular} & 93.1 & 105.3 & 68.1 & 96.5 \\
        HybrIK~\cite{li2021hybrik} & 89.1 & 98.6 & \textbf{66.2} & \textbf{89.1} \\
        PARE~\cite{kocabas2021pare} & \textbf{84.3} & \textbf{91.4} & 66.5 & 90.3 \\
        \bottomrule
        \end{tabular}
    \caption{Comparisons to state-of-the-art methods on the EgoHumans dataset using \textit{Protocol 2}~\cite{sun2021monocular} and train set similar to ~\cite{kolotouros2019learning}.}
    \label{supp:table:mesh}
\end{table}